\documentclass[letterpaper, 10 pt, conference]{ieeeconf}  

\IEEEoverridecommandlockouts                              

\usepackage{mathptmx} 
\usepackage{times} 
\usepackage{amsmath} 
\usepackage{amssymb}  

\usepackage{graphicx}
\usepackage{hyperref}

\title{\LARGE \bf
Curio: A Cost-Effective Solution for Robotics Education
}

\author{Talha Enes Ayranci$^{1}$, Florent P. Audonnet$^{1}$, Gerardo Aragon-Camarasa$^{1}$, Mireilla Bikanga Ada$^{1}$, Jonathan Grizou$^{1}$
\thanks{*T. E. A receive support from the Republic of Türkiye Ministry of National Education. This work also received support from the University of Glasgow's Chancellor's Fund.}
\thanks{$^{1}$School of Computing Science, University of Glasgow, United Kingdom
        {\tt\small t.ayranci.1@research.gla.ac.uk; gerardo.aragoncamarasa@glasgow.ac.uk}}%
}

\begin{document}

\maketitle
\thispagestyle{empty}
\pagestyle{empty}

\begin{abstract}
Student engagement is one of the key challenges in robotics and artificial intelligence (AI) education. Tangible learning approaches, such as educational robots, provide an effective way to enhance engagement and learning by offering real-world applications to bridge the gap between theory and practice. However, existing platforms often face barriers such as high cost or limited capabilities. In this paper, we present Curio, a cost-effective, smartphone-integrated robotics platform designed to lower the entry barrier to robotics and AI education. With a retail price below \$50, Curio is more affordable than similar platforms. By leveraging smartphones, Curio eliminates the need for onboard processing units, dedicated cameras, and additional sensors while maintaining the ability to perform AI-based tasks. To evaluate the impact of Curio on student engagement, we conducted a case study with 20 participants, where we examined usability, engagement, and potential for integrating into AI and robotics education. The results indicate high engagement and motivation levels across all participants. Additionally, 95\% of participants reported an improvement in their understanding of robotics. Findings suggest that using a robotic system such as Curio can enhance engagement and hands-on learning in robotics and AI education. 
All resources and projects with Curio are available at \href{https://trycurio.com/}{trycurio.com}.

\end{abstract}

\section{INTRODUCTION} \label{intro}

In computer science and robotics education, the key challenges students face include abstraction of programming concepts and the difficulty in applying theoretical knowledge to practical problems \cite{lahtinen_difficulties_2005, yilmaz_generative_2023}. These challenges can lead to frustration, reduced engagement, and lower motivation, particularly for students struggling with computational thinking \cite{figueiredo_increasing_2020}. To address these issues, tangible learning approaches have been widely adopted as tools to bridge the gap between theory and hands-on applications. For instance, Ortiz et al. \cite{ortiz_innovative_2017} introduced an innovative mobile robot method to enhance programming language learning in engineering education. Similarly, Ching et al. \cite{ching_computational} examined the role of educational technologies in fostering computational thinking among young learners.

Educational robots can provide an effective way to enhance engagement and learning by offering real-world applications of programming, robotics, and AI \cite{uslu_review_robotics_2022}. Among the various types of educational robots, mobile robots are particularly prevalent due to their ability to navigate diverse environments \cite{rubio_mobile_robots_2019}, and this mobility enables their usage in a wide range of scenarios \cite{raj_mobile_2022}.  For instance, learning-oriented robots such as Arduino-R2 \cite{arduino_r2}, AERobot \cite{aerobot} and MicroMVP \cite{micromvp} enable students to gain hands-on experience in robotics. Similarly, studies such as Mona \cite{mona} and Daran \cite{daran} explore the impact of robotics-based approaches on STEM education by providing accessible tools. Additionally, platforms such as Duckietown \cite{paull_duckietown_2017}, DonkeyCar \cite{donkeycar}, DeepRacer \cite{deepracer} and MuSHR \cite{srinivasa_mushr} focus on providing hands-on experience with self-driving technologies. However, while AERobot \cite{aerobot} and MicroMVP~\cite{micromvp} are affordable, they often lack the technical capabilities required for AI education due to limited sensor variety and the absence of a camera. On the other hand, advanced platforms, such as Deepracer\cite{deepracer} and MuSHR \cite{srinivasa_mushr}, have a higher price tag, which hinders the adoption of a 1-1 student-to-robot ratio. This financial barrier may render these platforms unaffordable for educational institutions \cite{serrano_perez_ultra-low_2019}. Although Atman Uslu et al. \cite{uslu_review_robotics_2022} indicated a significant increase in educational robotic platforms over the last years, the demand for educational robotics, especially cost-effective robots, in higher education and research \cite{paull_duckietown_2017} has not been met.

\begin{figure}[t]
\centerline{\includegraphics[width=0.9\linewidth]{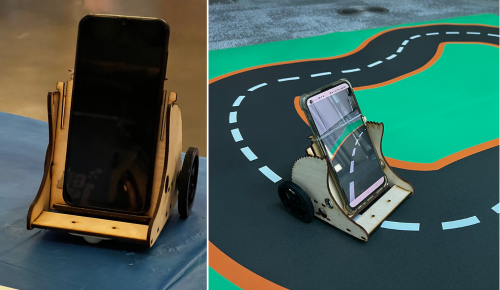}}
\caption{The Curio robot. Left: Back view with a smartphone. Right: Curio in a self-driving scenario going around a track.}
\label{curio_general}
\end{figure}

One possible way to create a cost-effective robot is to use smartphones, given that smartphone ownership among individuals aged 16 to 35 in the UK is around 98\% \cite{smartphone_ownership}. Their advanced capabilities, including powerful cameras, sensors, and CPUs capable of AI processing \cite{deeplearningphone}, make them a viable option for serving as the main information unit of robots. This approach could reduce robot costs by allowing users to use their smartphones. In this paper, we present Curio which is an affordable and smartphone-integrated robotics platform designed to lower the economic entry barrier to robotics and AI education. By leveraging smartphones, Curio eliminates the need for onboard computation, cameras, and various sensors while retaining the ability to perform AI tasks. Additionally, Curio supports web-based development, which allows users to interact with the robot without requiring dedicated software setups on the smartphone. Since Curio uses smartphones, mobile applications can be used as well to enhance computational capabilities and interaction.  Curio also has Python support, which allows flexibility for users and makes it smartphone-agnostic. In this paper, we demonstrate Curio's usability and impact on student engagement using a face-tracking task using a smartphone's camera. The key contributions of Curio are:

\begin{itemize}
    \item A cost-effective mobile robot to fill the 1-1 student-to-robot ratio gap.
    \item A smartphone-integrated environment to eliminate the need for onboard computation, dedicated cameras, and additional sensors while maintaining the ability to perform AI tasks.
    \item A hands-on platform to enhance student engagement in robotics and AI education.
\end{itemize}


\section{RELATED WORK} \label{related_work}

Motivation is crucial in enhancing students’ willingness to learn, particularly in AI education \cite{lin_motivation_2021}. Various theoretical frameworks have been explored, such as Lin et al. \cite{lin_motivation_2021} where they highlighted the potential of the attention, relevance, confidence, and satisfaction (ARCS) \cite{keller_arcs_2010} motivational model as a structured approach for sustaining students' motivation to learn AI. Additionally, Xia et al. \cite{xia_sdt_2022} proposed a design based on self-determination theory (SDT) \cite{ryan_sdt_2017} to enhance motivation and engagement among diverse student groups in AI courses. Another study conducted by Kuo et al. \cite{kuo_pbl_2022} indicated that the design thinking project-based learning approach improved motivation significantly.

\begin{table}[t]
\caption{Comparison of Educational Robots. To date, Curio is the most affordable robot for postgraduate education and the only robot not requiring to install dedicated software.}
\begin{center}
\begin{tabular}{|c|c|c|c|c|c|c}
\hline
\textbf{Platform} & \rotatebox{90}{Year} & \rotatebox{90}{\parbox{1.5cm}{\centering Suitable for \\ Postgraduate$^{\mathrm{1}}$ \\ Education}} & \rotatebox{90}{\parbox{1.5cm}{\centering Standalone \\ System}} & \rotatebox{90}{\parbox{1.5cm}{\centering Requires \\ Dedicated \\ Software}} & \rotatebox{90}{\parbox{1.7cm}{\centering Commercially \\ Available}} \\
\hline
AE Robot \cite{aerobot} & 2015 & X & X & \checkmark & X\\
\hline
Arduino-R2 \cite{arduino_r2} & 2021 & \checkmark & \checkmark & \checkmark & X\\
\hline
Daran \cite{daran} & 2021 & X & X & \checkmark & X\\
\hline
Deepracer \cite{deepracer} & 2019 & \checkmark & \checkmark & \checkmark & \checkmark\\
\hline
Donkey Car \cite{donkeycar} & 2019 & \checkmark & \checkmark & \checkmark & \checkmark\\
\hline
Duckiebot \cite{paull_duckietown_2017} & 2017 & \checkmark & \checkmark & \checkmark & \checkmark\\
\hline
microMVP \cite{micromvp} & 2017 & X & \checkmark & \checkmark & \checkmark\\
\hline
Mona \cite{mona} & 2018 & \checkmark & \checkmark & \checkmark & X\\
\hline
MuSHR \cite{srinivasa_mushr} & 2019 & \checkmark & \checkmark & \checkmark & X\\
\hline
OpenBot \cite{openbot} & 2020 & \checkmark & X & \checkmark & \checkmark\\
\hline
\textbf{Curio} & 2025 & \checkmark & X & X & X\\
\hline
\multicolumn{7}{l}{$^{\mathrm{1}}$Courses introducing machine learning, artificial intelligence or} \\ \multicolumn{7}{l}{computer vision concepts.}
\end{tabular}
\label{comparison}
\end{center}
\end{table}

Additionally, the strategy of visualizing abstract theoretical concepts for students and employing tangible tools such as simple robots in computer science education is a frequently tried approach for enhancing understanding and engagement \cite{uslu_review_robotics_2022, mamatnabiyev_holistic_2024, wang_effectiveness_2023}. This method includes using visual programming tools such as Scratch \cite{scratch} in computer science education \cite{merkouris_teaching_2017} as well as organizing robotics camps to teach computer science (CS) concepts \cite{qu_cultivating_2021}. Thanyaphongphat et al. \cite{visualization} demonstrated that a visual programming approach enhances the comprehension of iterative loops. That is, they integrated LEGO Mindstorms with game-based learning activities to allow students to visualize and engage with the concept of loops interactively.

Many robot platforms have been designed and used for educational purposes in classrooms or workshops (Table \ref{comparison}). Commercialization is widespread across some of these platforms, such as Deepracer \cite{deepracer}, Duckiebot \cite{paull_duckietown_2017}, and OpenBot \cite{openbot}. However, covering the financial burden associated with them presents a significant challenge for both students and universities \cite{serrano_perez_ultra-low_2019}. For instance, Duckiebot is priced at \$150 \cite{paull_duckietown_2017}, Donkey Car at \$200 \cite{donkeycar}, and Deepracer at \$400 \cite{deepracer}. It is important to note that these prices were reported at the time of publication and have not been adjusted for inflation, meaning their actual costs today may be lower or higher. 

The high cost of modern educational robots is largely attributed to their advanced hardware features, such as distance and speed sensors, cameras, LIDARs, and processors, as seen in platforms such as AutoRally \cite{goldfain_autorally_2019}, MuSHR \cite{srinivasa_mushr}, and F1Tenth \cite{f110}. Recently, there has been a rise in the development of low-cost robots, priced under \$50, by minimizing computation power and sensor variety. Examples include AERobot \cite{aerobot} and Arduino-R2 \cite{arduino_r2}, along with another low-cost design proposed by Serrano Pérez and Juárez López \cite{serrano_perez_ultra-low_2019}. Even though compromising on hardware features can lead to the creation of more affordable robots \cite{serrano_perez_ultra-low_2019}, sufficient processing power remains essential for using robots in the education of advanced technologies such as AI \cite{paull_duckietown_2017}. Given this challenge, smartphones equipped with diverse and advanced sensors, high-quality cameras, and sufficient processing power can serve as a practical solution, as they have become an integral part of daily life \cite{deeplearningphone}. OpenBot \cite{openbot} has demonstrated how smartphones can be integrated into robotics to achieve this goal, but it still requires the installation of dedicated applications on the smartphone.
To address the need for an accessible and technically capable robot for AI and robotics education, we developed Curio, 
a mobile robot for education and research (Fig. \ref{curio_general}).

\section{CURIO} \label{curio}

By leveraging smartphone-based processing and sensing, we aimed to eliminate the need for onboard computation while maintaining the capability of creating applications such as vision-based navigation and AI-driven control. Users can connect Curio via Bluetooth and control it through web applications they develop or programming environments such as Python. The main reason for supporting web applications is our intention to provide a zero-installation architecture without requiring additional setup on controller devices. 

\begin{figure}[t]
\centerline{\includegraphics[width=0.9\linewidth]{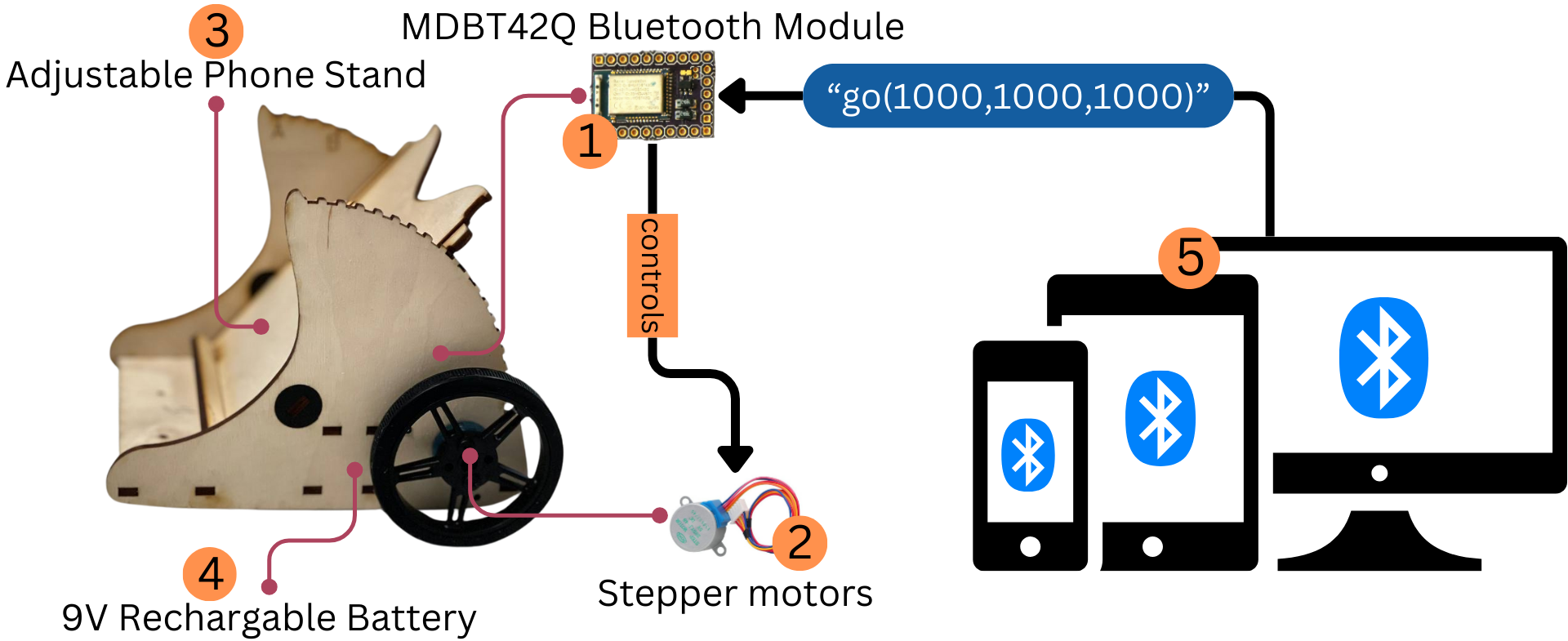}}
\caption{Components of the Curio robot and the communication process between Bluetooth devices (5). MDBT42Q module (1) handles the received command and controls the stepper motors (2) connected to wheels.}
\label{curio_features}
\end{figure}

For Curio, we chose a two-wheeled, car-like structure to provide mobility for various applications. The two-wheel system provides essential mobility while presenting a more economical option compared to a four-wheeled system. Wheeled robots, unlike humanoid or multi-legged robots, require fewer motors and have fewer moving parts, reducing maintenance needs and overall complexity. Curio's wheeled design provides straightforward motion with minimal hardware. This design choice not only lowers repair costs and extends the robot’s lifespan but also makes Curio more suitable for younger users, as the risk of damage is minimized. 

\subsection{Hardware}

Curio is designed to integrate smartphones as the primary computational unit. Therefore, our design comprises an adjustable smartphone mount ((3) in Fig. \ref{curio_features}) that supports flexible camera positioning for tasks such as visual tracking and real-time object detection. This smartphone stand can be adjustable up to 90 degrees, allowing users to orient the smartphone camera downward or forward depending on the requirement of the learning activity. The smartphone base is not connected to any motors and can be moved manually.

The robot body is constructed from laser-cut 3mm plywood. This material was chosen for its balance between durability and cost-effectiveness. While the robot body can also be 3D-printed, plywood was chosen as a more cost-effective option for our implementation and the possibility of mass production. The bottom of the robot has a rear metal ball caster for balance. Motion is achieved using stepper motors, which offer precision and allow users to program accurate motion sequences. Stepper motors provide controlled and incremental movement, and these are suitable for educational and experimental applications where fine-tuned control is important.

Curio’s communication system is a custom-designed board based on the Espruino \cite{espruino} MDBT42Q module ((1) in Fig. \ref{curio_features}), which was selected for its ease of programming with JavaScript (JS) commands and its built-in Bluetooth communication capability. The pre-installed Espruino JS interpreter\footnote{\href{https://www.espruino.com}{https://www.espruino.com}} enables users to execute JS commands directly. This enables web-based applications for rapid prototyping and remote interaction without the need for specialized embedded development tools. Additionally, the Bluetooth Low Energy (BLE) support in the MDBT42Q module allows wireless interaction with smartphones and other BLE-enabled devices while maintaining low power consumption. Curio operates on a single rechargeable 9V battery ((4) in Fig. \ref{curio_features}) with a micro-USB charging port. 
This design ensures an average operational time of 6 hours under normal usage, such as developing and testing projects, rather than continuous movement.

\subsection{Software}

Curio supports JS commands for operation, enabled by the Espruino JS interpreter running on the MDBT42Q Bluetooth module. Users can interact with the robot using UART libraries, which are widely available in multiple programming languages, including JS and Python. Through the UART interface, users can send JS commands as strings to Curio for execution. Curio includes a pre-defined function, "\texttt{go(left\_steps, right\_steps, speed)}" (Fig. \ref{curio_features}), which controls the step count for the left and right stepper motors with an optional speed parameter. For instance, sending the command "\texttt{go(1000, 1000, 1000)}" via UART moves Curio forward by executing equal steps on both motors at a defined speed. This simple command structure allows users to control Curio efficiently with minimal setup.

\paragraph{Web-Based Control}

Since JavaScript provides libraries for UART communication, WebBluetooth can be utilized directly in browsers. Currently, all Bluetooth-enabled smartphones, except iOS devices, support WebBluetooth and it allows browsers to establish Bluetooth connections. By including \texttt{UART.js}\footnote{\href{https://www.espruino.com/UART.js}{https://www.espruino.com/UART.js}} library into their projects, users can develop and run web-based applications locally or deploy them online without installing 3rd party applications. Through WebBluetooth API, we enable Curio to access and utilize smartphone sensors such as the camera, GPS, and microphone. This web-based approach allows for the sharing of projects using a URL or QR code as well as deployment on platforms such as GitHub Pages and Replit \footnote{\href{https://replit.com}{https://replit.com}}. Replit provides a multi-user platform to develop any kind of application collaboratively. Through Replit or similar platforms, users can code together in a live environment while testing Curio’s responses instantly. It offers a peer programming environment and makes the development process more collaborative.

\paragraph{Python Support and External Development} While web support of Curio brings independency to the development environment, there is a demand for Python language support for Curio since Python is one of the most frequent languages for robotics and AI development \cite{soori_ai_2023}. To allow users to interact with the robot using an external programming environment via Bluetooth, we developed a Python library \footnote{\href{https://github.com/CurioDev/curio-python}{https://github.com/CurioDev/curio-python}} based on the Bleak Bluetooth library \footnote{\href{https://github.com/hbldh/bleak}{https://github.com/hbldh/bleak}}. Additionally, while Curio is designed to leverage smartphones for processing, our observations indicate that in many cases, users may prefer to offload computationally intensive tasks to a PC while using the smartphone as a sensor or camera module. The Python library facilitates this option, allowing users to integrate PCs with Curio. Therefore, this Python library allows users to (a) Connect their PC to Curio via Bluetooth and send movement commands programmatically; (b) Use their smartphone as an IP camera mounted on Curio for real-time streaming; Develop custom applications in Python while leveraging Curio's Bluetooth-based control system; and, Enhance computational power by offloading processing to a PC.

We added Python support to make Curio suitable for advanced robotics experiments and custom research applications in the future. The motivation is to enable the use of Curio in university robotics labs for practical learning sessions or to adapt it for use in AI research projects with Python.

\section{CASE STUDY} \label{study}

The case study was designed to assess \textit{whether using a tangible system such as Curio increases students’ motivation to learn AI and robotics or helps maintain motivation among those already studying these topics}.  Additionally, this case study serves as an opportunity to test Curio in preparation for its planned integration into a future robotics course, allowing us to refine its usability and effectiveness in a structured educational environment. We also wanted to examine the user's attitudes toward such a robot since it could be classed as a social robot which interacts with people. Although we were not assessing participants' performance in this study, we wanted to examine the impact of Curio on users' understanding about robotics and AI topics. This study involved a structured hands-on session where participants interacted with Curio, followed by a survey assessing their experience. 

\subsection{Methodology}

As described in Section \ref{curio}, Curio is a two-wheeled mobile robot, and its main use case is navigating towards a goal. In this experiment, we decided to define a face-tracking task as a target since face-tracking includes AI to detect faces and robotics for managing the motion of the robot. We used the Python library since we defined this experiment as a pilot experiment in preparation for future possible classroom usage. The experimental setup included a Curio robot, a smartphone that can be users' own or ours, and a PC provided by us. The smartphone mounted on Curio ran an IP camera application\footnote{\href{https://droidcam.app}{https://droidcam.app}} and was connected to the same wireless network as the PC. Since we did not have our own IP camera application or website, the IP camera application was any IP camera application can be found on the market. Through the wireless network, the PC could connect to the smartphone camera and receive the real-time video stream via the Curio Python library. In the same way, a PC could connect to the Curio robot to send commands for movement.

\begin{figure}[t]
    \centering
    \includegraphics[width=0.7\columnwidth]{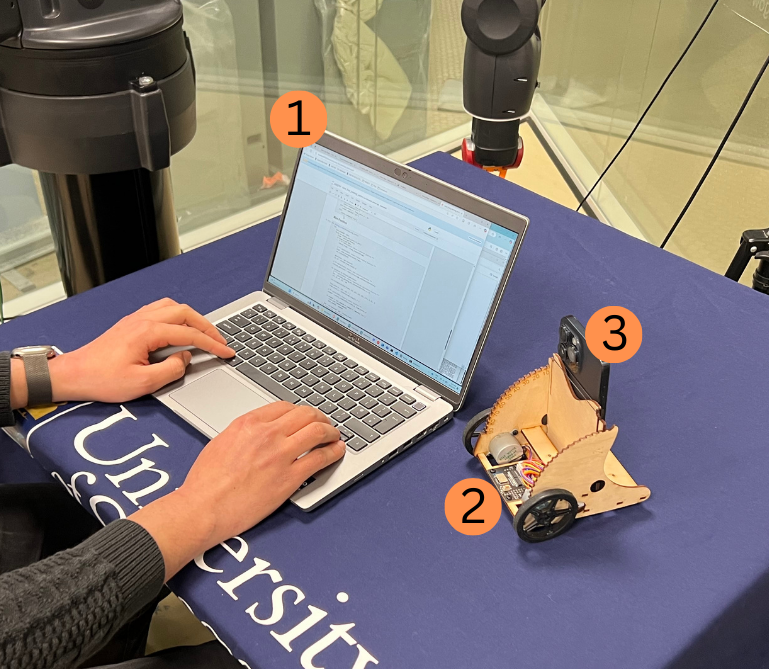}
    \caption{Overview of the case study environment. The laptop (1) is running a Jupyter Notebook, which runs a face-tracking algorithm. The Curio robot (2). The smartphone (3) is mounted on the robot and functions as a camera.}
    \label{experiment_setup}
\end{figure}

\begin{figure*}[t]
\centerline{\includegraphics[width=0.9\textwidth]{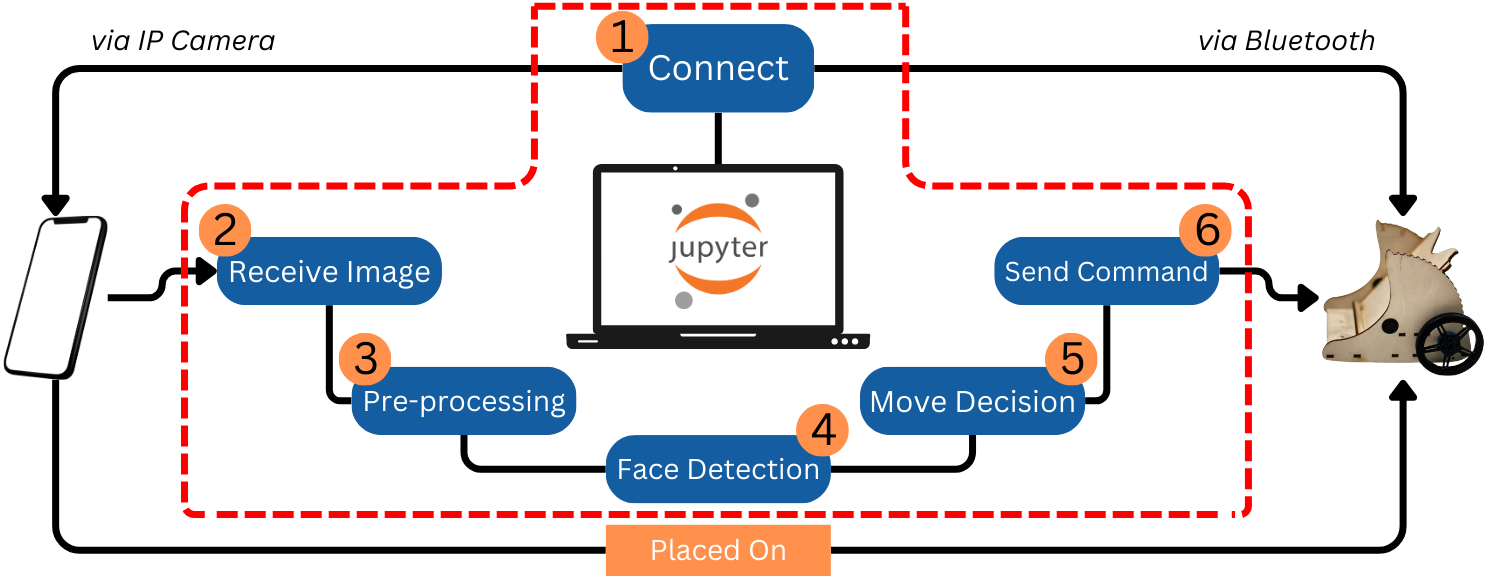}}
\caption{The flowchart of the face-tracking activity. The process begins with the Jupyter Notebook establishing a connection (1). The smartphone, acting as an IP camera, captures and transmits images to the notebook (2). The received image undergoes pre-processing (3) before a face detection algorithm is applied (4). Based on the detected face's position, the system makes a movement decision (5). Finally, a command is sent via Bluetooth (6) to the Curio robot, which adjusts its movement accordingly. The red dashed box highlights the computational process performed in the Jupyter Notebook.}
\label{activity_flowchart}
\end{figure*}

Since the aim of this experiment was to focus on student motivation and engagement, the face-tracking activity was pre-coded by us\footnote{\href{https://github.com/CurioDev/face-tracking-case-study/blob/main/curio_face_tracking.ipynb}{https://github.com/CurioDev/face-tracking-case-study/blob/main/curio\_face\_tracking.ipynb}} before the experiments. We created a Jupyter Notebook using the Curio Python library. The notebook consists of the connection commands for the robot and smartphone, the main algorithm which is face-tracking and helper functions. The task for the user was to modify the main function of the Jupyter Notebook to make the face-tracking algorithm more accurate from their point of view. It means there was not a correct answer for this task. The provided main function contained five key configurable functions that could impact the algorithm’s performance. For each of these functions, users had three options, allowing them to fine-tune the system. The five configurable points included:

\begin{itemize}
    \item Image rotation adjustments (correcting orientation for better detection)
    \item Brightness and contrast tuning (enhancing image quality for improved face recognition)
    \item Face detection algorithm confidence level
    \item Face bounding box margins (used to determine the center of the detected face)
    \item Robot control response speed (fast or slow adjustments)
\end{itemize} 

These five components are the main parts of our face-tracking algorithm. We did not want to give users complete control over the entire algorithm code, as we wanted to focus on their engagement in robotics and AI tasks, and having a complicated task would add some other variables to their motivation levels. The only parts the user should interact with are these five configurable points. A cheat sheet \footnote{\href{https://github.com/CurioDev/face-tracking-case-study/blob/main/Cheat\%20Sheet.pdf}{https://github.com/CurioDev/face-tracking-case-study/blob/main/Cheat\%20Sheet.pdf}} was provided, explaining the function of each parameter along with its advantages and disadvantages. 

We aimed for users to employ a trial-and-error approach to identify the optimal parameter combination for their face-tracking robot. There was no single correct solution for the task and all combinations were viable. However, the options we added for each function had an impact on the algorithm, and some could improve its performance compared to others. The goal for the user was not to find the absolute correct answer but rather to identify the most suitable selections based on their perspective by observing real-time results. We could have simplified the process by developing a no-code environment or a GUI, making it easier for users to choose options. However, we wanted users to interact with source codes to create a realistic educational environment at university level.

The case study was open to any computing science students from undergraduate to PhD. We recruited 13 master’s and 7 PhD students who volunteered without a monetary reward. The participants were required to have fundamental Python knowledge, but prior experience in AI or robotics was not mandatory. The entire session lasted a maximum of one hour, including a brief introduction, hands-on activity, and a user survey. The survey had 35 questions\footnote{\href{https://github.com/CurioDev/face-tracking-case-study/blob/main/Survey_Questions.pdf}{https://github.com/CurioDev/face-tracking-case-study/blob/main/Survey\_Questions.pdf}}. We used a System Usability Scale (SUS) \cite{sus} assessment to measure the usability of Curio since SUS is the most common and ideal tool for evaluating educational technology \cite{vlachogianni_sus_2021}. Additionally, a Negative Attitude Towards Robots (NARS) \cite{nars} questionnaire was used to assess participants’ biases or concerns about robotic systems. The majority of studies have used NARS to assess the perception towards robots \cite{bekesy_nars_2024}. The scale consists of three subscales: S1 measures negative attitudes toward interactions with robots, S2 evaluates concerns about the social influence of robots, and S3 assesses negative attitudes toward emotional interactions with robots. We only used the S1 and S2 subscale questions from NARS since S3 was not relevant for this use case. Learning outcomes questions focused on the understanding levels of the participants about robotics and their confidence in implementing AI-based applications after the experiment. Motivation and engagement questions were included to understand how participants felt during the experiment and how Curio influenced their interest and involvement in the learning process. 

To understand participants' behavior while implementing an algorithm, we asked two behavioral questions about the factors influencing their choices. For selecting specific functions from the available options, we provided six key considerations: ease of implementation, accuracy, performance efficiency, flexibility, stability, and prior experience. Similarly, for evaluating the most important aspects of successful tracking, we offered six options: maintaining target face visibility, processing speed, environmental conditions, stability of movement, minimizing false positives, and ease of debugging. Participants were allowed to select multiple options when responding. All survey items, except demographics and the previously mentioned two behavioral questions, which allow multiple selections, used a 5-point Likert scale, which measured the degree of agreement with various statements from 1 (strongly disagree) to 5 (strongly agree).

\subsection{Results}

Among the 20 participants, 14 had prior experience with similar activities, while 6 had not. Participants exhibited diverse backgrounds in robotics and AI across four experience levels: no experience, basic, moderate, and highly experienced. Nine had no robotics experience (SD = 1.08), and one had no AI experience (SD = 0.97).

\begin{figure}[t]
\centerline{\includegraphics[width=0.9\linewidth]{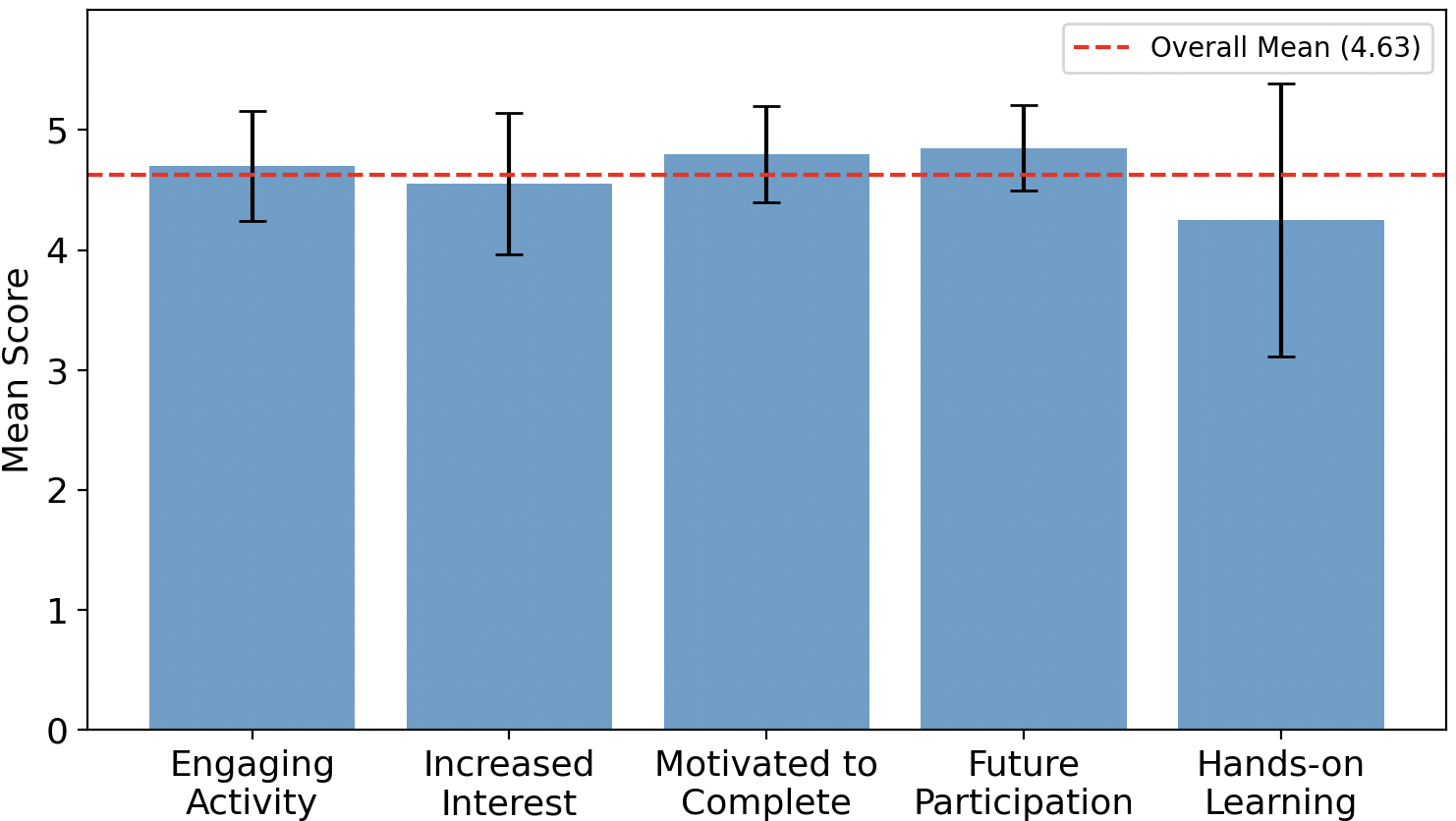}}
\caption{Mean scores of engagement-related questions with standard deviation error bars. The red dashed line represents the overall mean score (4.63).}
\label{engagement_means}
\end{figure}

Engagement and motivation levels were consistently high across all participants (Fig. \ref{engagement_means}). We calculated agreement rates for each engagement question by combining the user's responses as \textit{Agree} or \textit{Strongly Agree}. The agreement rates indicate that all participants found the activity engaging and reported being motivated to complete the task with a willingness to participate in similar activities. Additionally, 95\% of responses agreed on an increased interest in AI/robotics, and 80\% agreed on the impact of hands-on learning on engagement. Notably, as observed in Fig. \ref{engagement_means}), the highest mean score (4.85, SD = 0.36) was for willingness to participate in similar activities in the future, which suggests that Curio successfully sustained the user's long-term interest. While engagement and motivation scores were consistently high (4.70, SD = 0.46 and 4.80, SD = 0.40, respectively), the lower mean score for the impact of hands-on learning (4.25, SD = 1.13) suggests some variation in how participants perceived the effectiveness of hands-on learning. The engagement results indicate that users found the idea of using a tangible system such as Curio for robotics and AI-based activities to be engaging.


\begin{figure}[t]
\centerline{\includegraphics[width=0.9\linewidth]{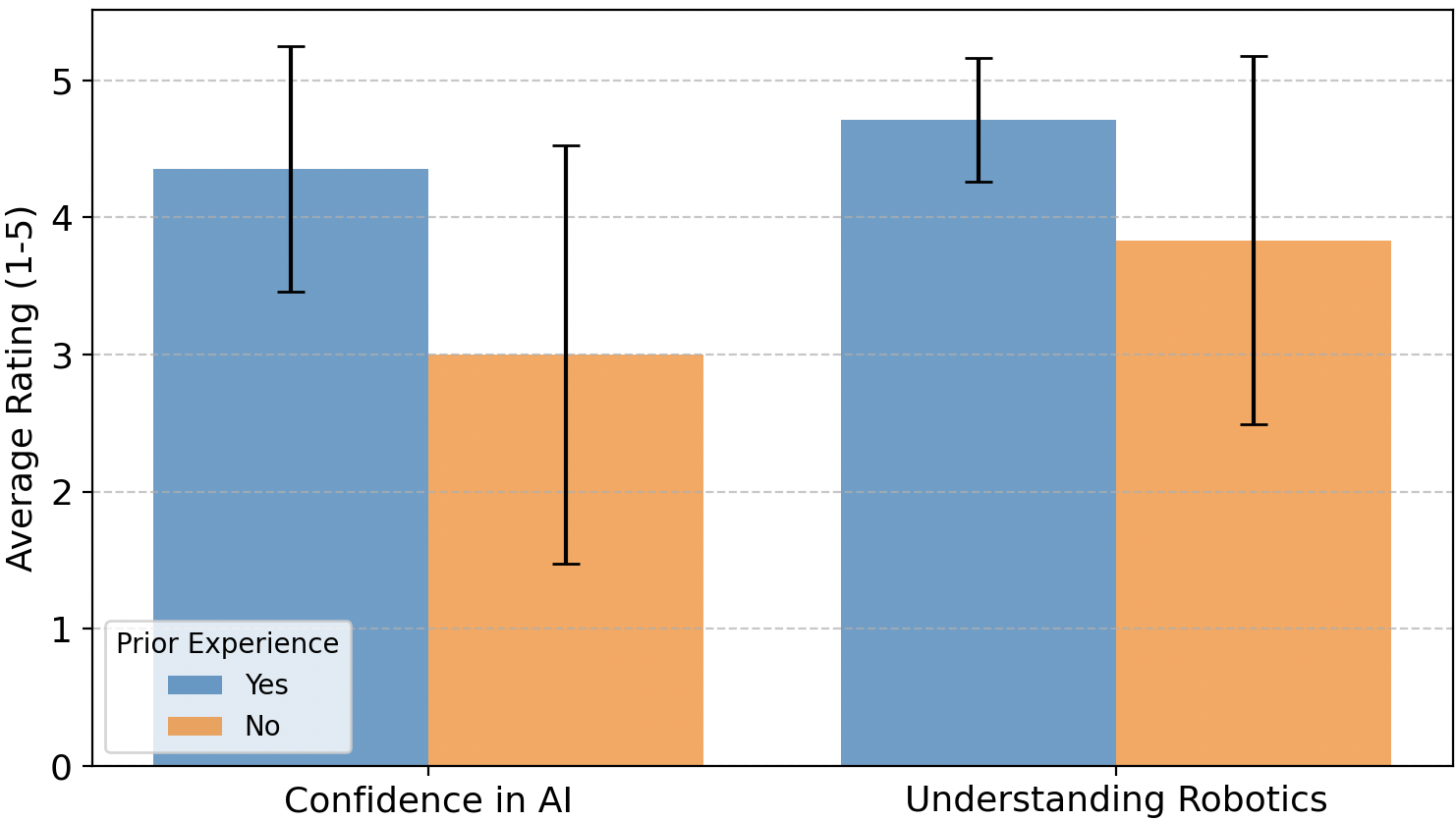}}
\caption{Comparison of experienced and non-experienced groups in terms of understanding.}
\label{understanding_prior_experience}
\end{figure}

Regarding learning outcomes, 95\% of participants agreed that their understanding improved about robotics. For the question about their confidence in the knowledge of AI-based applications, 75\% indicated that they felt more confident about creating similar applications using AI models after this activity. The mean scores were 4.45 (SD = 0.92) for understanding robotics and 3.95 (SD = 1.28) for confidence in AI. These results suggest that the activity enhanced participants' understanding of robotics and increased their confidence in developing AI-based applications. However, as indicated in Fig. \ref{understanding_prior_experience}, participants who had carried out similar activities before demonstrated higher confidence in developing similar AI-based projects in the future, with a mean score of 4.357 (SD = 0.89) compared to 3 (SD = 1.53) out of 5 for those who had not attended similar activities. Additionally, their understanding of real-time robotics was also higher, averaging 4.714 (SD = 0.45) compared to 3.833 (SD = 1.34) for those without prior experience. These findings suggest that participants with prior experience in AI and robotics were able to connect their existing knowledge with the platform, allowing them to achieve a deeper understanding of real-time robotics compared to those without prior experience.

\begin{figure}
\centering
   \includegraphics[width=0.9\linewidth]{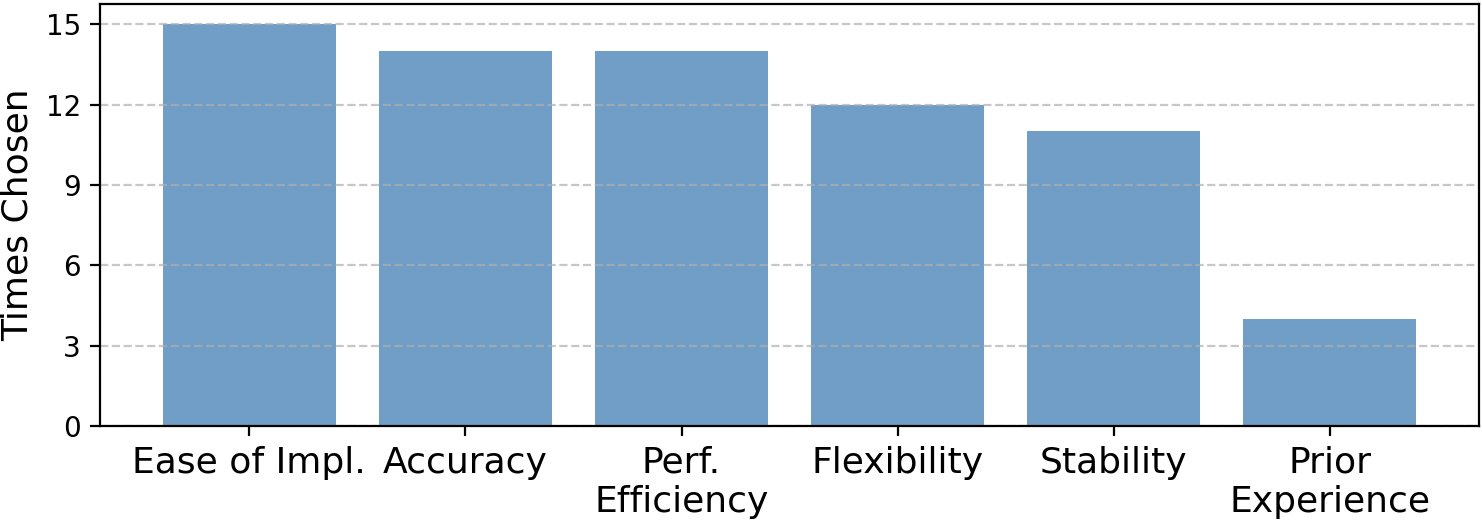}
   \includegraphics[width=0.9\linewidth]{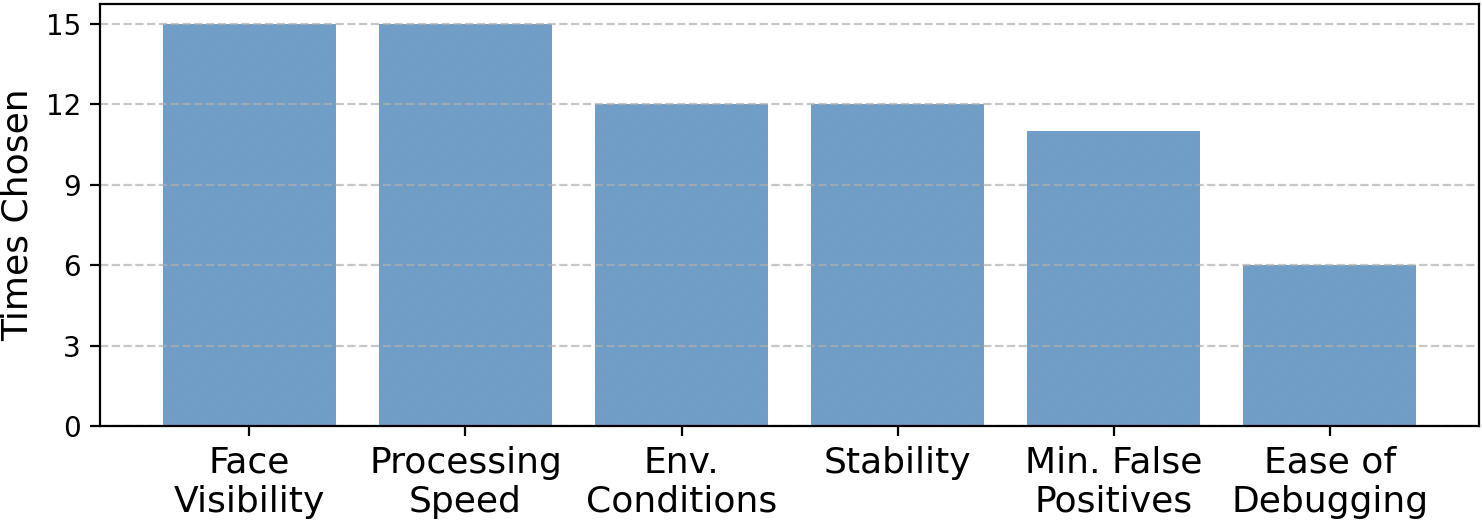}
    \caption{The frequency of the options for behavioral questions. The most common reasons for selecting a specific function (top) and considerations for the tracking success evaluation (bottom).}\label{behavior_1} \label{behavior_2}
\end{figure}

User interaction with Curio revealed interesting behavioral patterns. As shown in Fig. \ref{behavior_1} top, the most influential factors in selecting specific function settings were \textit{accuracy}, chosen 15 times, followed by \textit{flexibility} and \textit{stability}, each selected 14 times. In contrast, \textit{prior experience} was the least selected factor with only 4 mentions. This suggests that participants made their selections through observation of the robot's movements, suggesting that even those without prior knowledge could effectively use the system. When evaluating tracking success (Fig. \ref{behavior_2} bottom), participants most frequently prioritized \textit{maintaining target visibility} and \textit{processing speed}, both selected 15 times. The less frequently selected factor was \textit{ease of debugging} as users were more concerned with maintaining real-time tracking than troubleshooting errors. The overall system usability score (SUS) was calculated as 88 from the answers given to the SUS questions. Scores higher than 85 fall within the 'excellent' usability range \cite{sus_scale}, indicating a high level of perceived usability.

Several correlations were identified using the Pearson correlation coefficient (r). An absolute r value between 0.4 and 0.59 indicates a moderate correlation, 0.6 to 0.79 indicates a strong correlation, and 0.8 to 1.0 indicates a very strong correlation. That is,

\begin{enumerate}
    \item \label{cor_1} Participants who would feel uneasy about using Curio in a professional setting tended to report a lower understanding of controlling robots (r = 0.75). 
    \item \label{cor_2} Users who perceived Curio as consistent and predictable were more motivated to complete the task (r = 0.75).
    \item \label{cor_3} Those who expressed a stronger desire to use Curio more frequently also reported greater confidence in optimizing AI-based algorithms (r = 0.61).
    \item \label{cor_4} Participants who found Curio engaging showed a greater interest in using it frequently (r = 0.56).
\end{enumerate}

Correlation \ref{cor_1} indicates that when users feel more comfortable, they are more likely to develop a stronger understanding with a strong correlation. Another strong correlation appears in \ref{cor_2}, and it suggests a smooth and well-integrated system can enhance engagement.

Several suggestions were also provided from the participants for improving or adapting Curio’s features within an educational context. Participants highlighted key areas that could enhance Curio’s effectiveness and usability. That is, several participants found the motors slow, even at maximum speed. Additionally, since the smartphone stand is adjustable, some participants experienced difficulties keeping the smartphone securely in place, and in some cases, the smartphone occasionally fell due to insufficient support. Participants also emphasized modularity, suggesting enhancements such as GPIO access for external sensors and actuators. They recommended modifications to the physical enclosure to allow for attachments and customizations, which would encourage creativity and hands-on experimentation. Additionally, they proposed introducing USB connectivity as an alternative to Bluetooth, ensuring a more stable and reliable connection for programming and control. These suggestion will be considered in the next iteration of the Curio's design.

\section{CONCLUSION AND FUTURE WORK} \label{conclusion}

This paper introduces the Curio robot as an affordable robot for education. Its retail price is below \$50, which is cheaper than similar platforms such as Duckiebot \cite{paull_duckietown_2017}, OpenBot \cite{openbot} and Deepracer \cite{deepracer}. We evaluated its usability and impact on student engagement. The results of the evaluation with 20 participants are promising. Participants found the hands-on activity engaging, and the experience of working with a physical robot contributed to a deeper understanding of robotics concepts. This suggests that Curio has the potential to enhance engagement and improve learning experiences on robotics and AI topics.

Future work will involve developing the next version of Curio by addressing the issues and applying some suggestions identified during the case study. The updated Curio will then be integrated into a foundational robotics course as a laboratory tool to evaluate its impact in a real educational setting. The planned course at the University of Glasgow currently utilizes ROS2 \cite{macenski_ros2_2022} with the Gazebo \cite{koenig_gazebo_2004}  simulator, and our goal is to integrate Curio in place of the simulation to assess the impact of tangible tools on student engagement and understanding.

A potential avenue for future work with Curio is its utilization as a testbed for robotics research. As highlighted by Wilson et al. \cite{wilson_robotarium_2020}, the evaluation of robot control algorithms often incurs significant costs in terms of resources, time, and expertise. To address these challenges, they developed a remote swarm robotics platform designed to reduce such expenditures. We propose positioning Curio as a valuable resource for researchers in the fields of artificial intelligence and robotics. This can be achieved by allowing PhD and Master's students to implement and test their methodologies on Curio, thereby facilitating research and innovation in these domains.









\end{document}